\documentclass[11pt]{article}

\usepackage[final]{acl}

\usepackage{times}
\usepackage{latexsym}

\usepackage[T1]{fontenc}

\usepackage[utf8]{inputenc}

\usepackage{microtype}

\usepackage{inconsolata}

\usepackage{graphicx}
\usepackage{algorithm}
\usepackage{algpseudocode}
\usepackage{multirow}
\usepackage{comment}
\usepackage{enumitem}
\usepackage{booktabs}
\usepackage{amsmath} 
\usepackage{graphicx}
\usepackage{amssymb}

\definecolor{chestnut}{cmyk}{0, 0.7808, 0.4429, 0.1412}
\definecolor{lightgreen}{RGB}{144,238,144}

\title{Unlocking the Potentials of Retrieval-Augmented Generation for Diffusion Language Models: A Semantic Drift Perspective}

\author{
\textbf{Chuanyue Yu\textsuperscript{1}}, 
\textbf{Jiahui Wang\textsuperscript{1}}, 
\textbf{Yuhan Li\textsuperscript{3}}, 
\textbf{Heng Chang\textsuperscript{4}}\thanks{Project Lead},
\\
\textbf{Ge Lan\textsuperscript{1}\footnotemark[1]}, 
\textbf{Qingyun Sun\textsuperscript{2}},  
\textbf{Jia Li\textsuperscript{3}}, 
\textbf{Jianxin Li\textsuperscript{2}}, 
\\
\textbf{Ziwei Zhang\textsuperscript{2}\footnotemark[1]\thanks{Corresponding Author}}
\\
 \textsuperscript{1}Nankai University,
 \textsuperscript{2}Beihang University,
 \textsuperscript{3}HKUST (Guangzhou),
 \textsuperscript{4}Huawei Technologies Co., Ltd.,
 \\
 \small{
   {yuchuanyue@mail.nankai.edu.cn},{changh.heng@gmail.com},{lange@nankai.edu.cn},{zwzhang@buaa.edu.cn}
 }
}

\begin{document}

\newcommand{\methodname}{SPREAD}
\newcommand{\methodnamefull}{Semantic-Preserving Retrieval-Augmented Diffusion}
\newcommand{\metricname}{RSD}
\newcommand{\metricnamefull}{Response Semantic Drift}
\maketitle
\begin{abstract}
Diffusion Language Models (DLMs) have recently demonstrated remarkable capabilities in natural language processing tasks. 
However, the potential of Retrieval-Augmented Generation (RAG), which shows great successes for enhancing large language models (LLMs), has not been well explored, due to the fundamental difference between LLM and DLM decoding. 
To fill this critical gap, we systematically test the performance of DLMs within the RAG framework.
Our findings reveal that DLMs coupled with RAG show promising potentials with stronger dependency on contextual information, but suffer from limited generation precision. 
We identify a key underlying issue: \metricnamefull~(\metricname), where the generated answer progressively deviates from the query's original semantics, leading to low precision content.
We trace this problem to the denoising strategies in DLMs, which fail to maintain semantic alignment with the query throughout the iterative denoising process. 
To address this, we propose \underline{S}emantic-\underline{P}reserving \underline{RE}trieval-\underline{A}ugmented \underline{D}iffusion (SPREAD), a novel framework that introduces a query-relevance-guided denoising strategy. By actively guiding the denoising trajectory, \methodname~ensures the generation remains anchored to the query's semantics and effectively suppresses drift. 
Experimental results demonstrate that \methodname~significantly enhances the precision and effectively mitigates \metricname~of generated answers within the RAG framework.%
\end{abstract}

\section{Introduction}

\begin{table*}[t]

  \setlength{\tabcolsep}{3.7pt}
   \resizebox{\linewidth}{!}{
  \begin{tabular}{lccccccccccccc}
    \toprule
     \multirow{2}{*}{Method}     & \multicolumn{2}{c}{NQ} & \multicolumn{2}{c}{Trivia} & \multicolumn{2}{c} {HotpotQA}    & \multicolumn{2}{c}{MuSiQue}    & \multicolumn{2}{c}{Multihop}  & \multicolumn{2}{c}{UltraDomain} \\

              \cmidrule[1pt](lr){2-3}\cmidrule[1pt](lr){4-5}\cmidrule[1pt](lr){6-7}\cmidrule[1pt](lr){8-9}\cmidrule[1pt](lr){10-11}\cmidrule[1pt](lr){12-13}\
                            & $\mathrm{F1}$        & $\mathrm{CR}$       
                            & $\mathrm{F1}$        & $\mathrm{CR}$       
                            & $\mathrm{F1}$        & $\mathrm{CR}$       
                            & $\mathrm{F1}$        & $\mathrm{CR}$       
                            & $\mathrm{F1}$        & $\mathrm{CR}$       
                            & $\mathrm{F1}$        & $\mathrm{CR}$  \\
 \midrule
    \text{Qwen2.5-7B}     &   28.33        &  77.59       &   40.95   &  82.68   &  47.44       &  77.79        &   22.25        &   64.87        &   26.98    &     58.67    &     22.21   & 45.42 \\
    \text{Qwen2.5-32B}    &   23.17        &  71.69       &   33.19   &   78.58  &  38.42       &  72.80        &   20.56        &   59.76        &   24.32    &    58.07     &     22.33   & 44.28 \\
    \text{Llama3-8B}      &   21.59      &  70.14       &    36.73  &   62.36  &  39.77       &  61.91        &   17.24        &  59.18         &   15.11    &     57.78    &     28.85   & 59.59  \\

    \midrule
    \text{LLaDA-8B}  &   33.09      &      81.01    &   39.80   &   82.24  &   44.36        &     79.82         &   27.03          &    72.36         &  \textbf{46.34}    &     \textbf{63.39}    &    30.91    & 61.78 \\
    \text{Dream-7B}  &   \textbf{38.42}      &  \textbf{84.81}        &   \textbf{46.47}    &   \textbf{84.41}  &   \textbf{52.40}         &    \textbf{81.88}          &   \textbf{30.56}          &     \textbf{77.65}         &  38.46   &    58.63    &   \textbf{36.92}      &  \textbf{71.35}  \\

    \bottomrule
  \end{tabular}
  }
  \caption{Performance comparison of LLMs and DLMs equipped with RAG. CR indicates copy rate. DLMs in RAG achieve high F1 scores and exhibit high copy rates from the provided context.}
    \label{tab:dlm_vs_llm}
\end{table*}

While auto-regressive Large Language Models (LLMs) have dominated natural language processing, recent advancements in Diffusion Language Models (DLMs)~\cite{llada,ye2025dream,DLM-discrete-first} present an attractive alternative, demonstrating superior performance in tasks requiring fine-grained control due to their parallel decoding and iterative refinement capabilities~\cite{DLM-survey}. However, similar to LLMs, DLMs are also confronted with challenges such as knowledge gaps and hallucination. On the other hand, Retrieval-Augmented Generation (RAG)~\citep{RAG-lewis} has become a foundational technology for enabling LLMs to generate reliable outputs grounded in external knowledge sources~\citep{RAG-survey,zhao2024retrieval}.
Consequently, integrating DLMs with RAG to mitigate these issues is a pivotal direction, yet remains largely unexplored in the literature. This problem is also highly challenging as DLMs and LLMs have fundamentally different generation processes. Whether the unique architecture of DLMs necessitates different RAG design warrants further investigation.

\begin{figure}
  \includegraphics[width=\columnwidth,page=5]{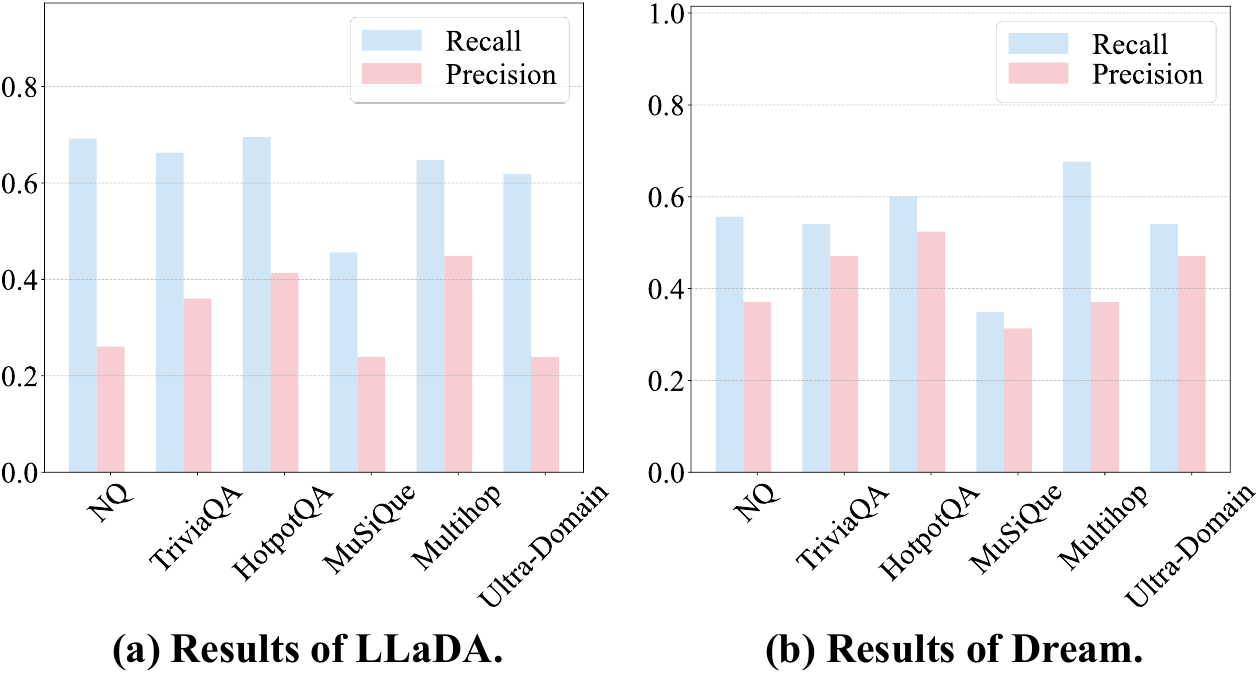}
  \caption{Precision and recall of DLMs equipped with RAG. While both LLaDA and Dream achieve high recall, their precision is consistently relatively low. }
  \label{fig:preliminary:precision}
\end{figure}

To bridge this gap, we conduct preliminary experiments to combine DLMs with RAG for the question answering task (the details are introduced in Section~\ref{sec:preexp:setup}). %
As shown in Table~\ref{tab:dlm_vs_llm}, DLMs can achieve scores comparable to or even surpassing those of LLMs on metrics such as F1. A particularly notable finding is their stronger tendency for high copy rate (CR), which we define as the proportion of words in the generated answer that appear identically within the provided source context. 
However, as further shown in Figure~\ref{fig:preliminary:precision}, the absolute precision of DLM-generated answers remains a bottleneck. The results suggest that while the copied retrieved content is factually grounded in the context, it may often include redundant details. Consequently, the answer's relevance to the query's semantics is diminished.

We identify a key issue underlying this phenomenon: \metricnamefull~(\metricname) within the generated answer. 
In contrast to the semantic drift in \citep{sd2024}, which measures the separation between correct and incorrect facts in generated text, or other task-specific semantic drifts~\citep{sd2024,khan2025measuring,mollah2025telephone},
our RSD focuses on the gradual drift in semantic between consecutive sentences within the generated content, which can lead to outputs that are coherent at a local level but lack query semantics relevance and precision. 
Our analysis further traces this issue to the inherent denoising strategies in DLMs, particularly confidence-based methods~\citep{llada,ye2025dream,DLM-confidence-slowfast,DLM-confidence-answer-before-decoding,DLM-noise-entropy}. These strategies determine denoising priorities based on global and token-level certainty estimates. However, they fundamentally overlook the alignment of the denoising process with the query semantics. Consequently, the generation process may gradually deviate from the original semantics, resulting in the observed semantic drift and precision loss.

To address this challenge and fully unlock the potentials of RAGs for DLMs, we propose \underline{S}emantic-\underline{P}reserving \underline{RE}trieval-\underline{A}ugmented \underline{D}iffusion (SPREAD). Our core idea lies in assessing the semantic relevance between generated token and query, and then utilizes these scores to determine token selection. This strategy actively guides the iterative denoising process of DLMs at each denoising step, ensuring that the decoded text remains closely aligned with query semantics throughout the generation process. Compared to previous denoising strategies that rely on global contexts and token-level certainty, \methodname~dynamically adjusts the denoising process, using query semantics as a constraint to suppress semantic drift. Experimental results show that our method consistently and substantially enhances the precision of the generated answers and effectively mitigates the issue of RSD across all datasets.

Our contributions are summarized as follows:
\begin{itemize}[leftmargin=0.5cm]
\item We investigate the integration of DLMs and RAG and identify \metricnamefull~as a key bottleneck, leading to redundant and incoherent outputs. To the best of our knowledge, we are the first to study this critical issue. 
\item We propose \methodname, a novel framework that introduces a query-relevance-guided denoising strategy to dynamically align the generation process with the query semantics.
\item We conduct extensive experiments to validate the effectiveness of \methodname. Our method significantly mitigates semantic drift and enhances output quality for both LLaDA and Dream across 6 benchmark datasets.
\end{itemize}

\section{Related Work}
\subsection{Retrieval-Augmented Generation}
RAG enhances language models by integrating external knowledge through joint modeling of retrieval and generation~\citep{RAG-lewis}. Subsequent work has pursued better knowledge integration via richer representations~\citep{RAG-Hierarchical-Knowledge} and end-to-end retriever optimization to align retrieval with generation goals~\citep{RAG-openrag-end-to-end,RAG-Optimizing-Differentiable-Data-Rewards}. Other efforts coordinate the RAG pipeline through joint training~\citep{RAG-Knowledge-Graph-Question-Answering}, reinforcement learning~\citep{RAG-multiple-partitions}, or unified ranking-generation schemes~\citep{RAG-rankrag-Unifying-context-ranking}. Furthermore, research has expanded to specialized data forms, such as developing RAG frameworks for table-based knowledge~\citep{chen2024tablerag}, and explored novel paradigms like organizing knowledge into multiple partitions for enhanced agentic reasoning~\citep{m-rag}.

While RAG is a vital solution for knowledge supplementation and hallucination reduction in generative models, its application to the DLM paradigm is unexplored. A key research question is whether the DLM’s unique architecture and decoding process require new methodologies for such integration.

\subsection{Diffusion Language Models}
DLMs offer a non-autoregressive alternative for LLMs, which generate text through iterative denoising for parallel decoding and enhanced controllability ~\citep{DLM-survey}. Recent research on DLMs has branched along several paths. Initial explorations with continuous DLMs operating in latent spaces demonstrated advantages in efficiency and controllability ~\citep{DLM-Energy-based,DLM-One-Step-Sequence-Generation}. However, for textual data, discrete diffusion methods have been developed as a more direct approach to better model  text tokens~\citep{DLM-discrete-estimating-the-ratios-of-the-data-distribution}. Building on this, masked and likelihood-based methods have demonstrated competitive performance, narrowing the gap with autoregressive models~\citep{DLM-masked,DLM-Likelihood-based}. Hybrid approaches like block diffusion further balance efficiency and flexibility ~\citep{DLM-Block}. 

However, the integration of DLMs with RAG remains a notably underexplored area. Even with a preliminary attempt for continuous DLMs~\citep{DLM-RAG}, the fundamental challenge of how to effectively ground the iterative denoising process of DLMs on retrieved knowledge has not been systematically explored. A core challenge is how to ensure that the iterative denoising process over tokens remains semantically anchored to the dynamically retrieved knowledge throughout the generation.

\section{Exploratory Experiments and Analyses}\label{sec:metric}

In this section, we conduct a detailed analysis to explore the potential and identify challenges of combining DLMs with RAG. We begin by presenting preliminary experiments, which reveal a key performance bottleneck, i.e., low answer precision, and motivate a deeper investigation into their generation process.

\subsection{Preliminary Results}\label{sec:preexp:setup}

To investigate the performance of DLMs equipped with RAG, we conducted some preliminary experiments on. For the retrieval component of RAG, we employ a dense vector retriever powered by the NV-Embed-v2 encoder~\citep{NV-embed}. All documents are first segmented into chunks of 2,000 characters. For each query, we retrieve the top-5 most relevant passages based on vector similarity. These retrieved passages are then concatenated and provided as the context to the language model for answer generation. We compare the performance of DLMs against LLMs on question answering benchmarks (detailed experimental setups are provided in Section~\ref{exp:mainsetup}).

The results are shown in Table~\ref{tab:dlm_vs_llm} and Figure~\ref{fig:preliminary:precision}. On the one hand, DLMs achieve factual accuracy scores comparable to or even surpassing their autoregressive counterparts on metrics such as F1. This appears to stem from their strong inclination to copy content verbatim from the provided context. However, compared to high recall, the absolute precision of DLM-generated responses often emerges as a distinct performance bottleneck. These results suggest that while directly adopting RAGs can empower DLMs to copy factually grounded text from retrieved knowledge, it may also include considerably redundant and irrelevant information. Consequently, the answer's relevance to the query's semantics is not fully optimized.

Overall, the results suggest that combining DLMs with RAGs provides promising potentials, but directly adopting the off-the-shelf RAG does not achieve optimal results. In the following subsections, we provide in-depth analyses for this phenomenon and motivate our proposed method. 

\subsection{\metricnamefull}
To gain further insights, we take a closer look at the iterative denoising process in discrete DLMs and observe a critical phenomenon: the generated content, while potentially factually related to the context, is not fully faithful to the query. 
Concretely, the outputs tend to be redundant and gradually diverge from the core semantics of the query. We term this problem affecting the generation quality as \metricnamefull~(\metricname).

Formally, given a generated answer text \(a\), we segment it into a consecutive sequence of sentences \(S = \{s_1, s_2, ..., s_{|S|}\}\). 
We measure the average semantic incoherence between adjacent sentences within the generated answer, calculated as follows:
\begin{equation}
\text{RSD}(a) = \frac{1}{|S| - 1} \sum_{i=1}^{|S|-1} \text{Dist}(s_i, s_{i+1}),
\end{equation}
where \(\text{Dist}(s_i, s_{i+1})\) denotes the semantic distance between two adjacent sentences \(s_i\) and \(s_{i+1}\). A larger value indicates more pronounced semantic drift between consecutive sentences, corresponding to poor overall coherence.

We employ cosine distance based on a high-performance pre-trained semantic encoder to quantify the semantic difference between sentences:
\begin{equation}
\text{Dist}(s_i, s_{i+1}) = 1 - \text{cosine}(\text{Enc}(s_i), \text{Enc}(s_{i+1})).
\end{equation}

\begin{figure}
  \includegraphics[width=\columnwidth,page=1]{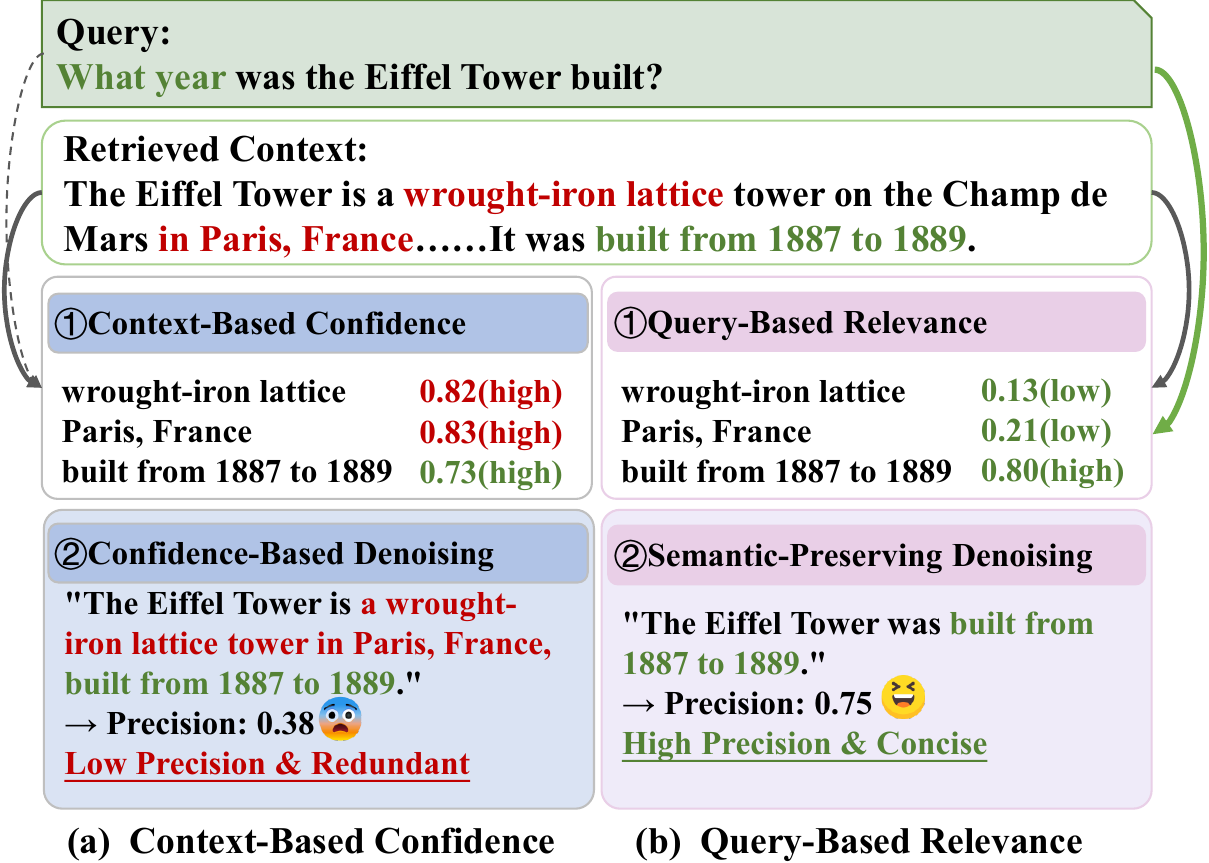}
  \caption{An illustration comparing different denoising strategies in DLMs. (a) Context-Based Confidence. The model relies on its internal confidence score, which is potentially biased by the global retrieved context, causing the model to generate irrelevant information.
(b) Query-Based Relevance. An ideal approach maintains a persistent focus on the query's semantics, which allows the model to prioritize relevant facts and generate a concise and relevant response.}
  \label{fig:example_problem_mini}
\end{figure}

We provide an illustrative example in Figure~\ref{fig:example_problem_mini}.
Consider a simple query such as ``What year was the Eiffel Tower built?'' along with its relevant context. A straight-forward DLM-RAG baseline model might generate an answer as shown in Figure~\ref{fig:example_problem_mini}(a). Although each sentence is independently correct and derived from the retrieved context, the response as a whole exhibits significant semantic drift. For instance, the content may jump from the tower's structural features to its construction timeline, thereby failing to precisely focus on the query's semantics regarding the construction date.
In contrast, an ideal response would directly address the core of the query as shown in Figure~\ref{fig:example_problem_mini}(b). Alternatively, it could expand on the core topic while maintaining high thematic coherence, e.g., ``The Eiffel Tower was built from 1887 to 1889. Its construction took two years.'' In this response, the \metricname~between consecutive sentences is minimal.

\subsection{Semantic Drift in the Denoising Process}
The problem of \metricname~stems from a fundamental conflict between the local optimization of the DLM's denoising process and the global semantic requirements of question answering with RAG tasks. 
The standard diffusion denoising process relies on strategies based on confidence or entropy. 
These strategies operate on a step-wise basis, which are inherently local and greedy. At each iteration, its primary objective is to select the most probable token based on the retrieved context and the current generative state. 
However, this process of locally optimal denoising lacks a global alignment for semantics. 
In question answering tasks, the objective is to maintain global alignment of the entire sequence with the query's semantics. The final generated answer must remain strictly aligned with the semantics of the initial query.
Since standard denoising strategies overlook this overall goal, the influence of the original query is reduced over successive iterations. Consequently, the generation process falls into a series of local decisions, and minor deviations at each step gradually accumulate and amplify throughout the iterative process. Finally, the model is likely to forget the query and generate technically correct but irrelevant answers. 
Therefore, resolving this issue requires a strategy that can guide the denoising process and enforce a global semantic constraint, especially the alignment with the query's semantics, which is the key motivation for our proposed method. %

\section{Method}
\begin{figure}[t]

  \includegraphics[width=\columnwidth,page=2]{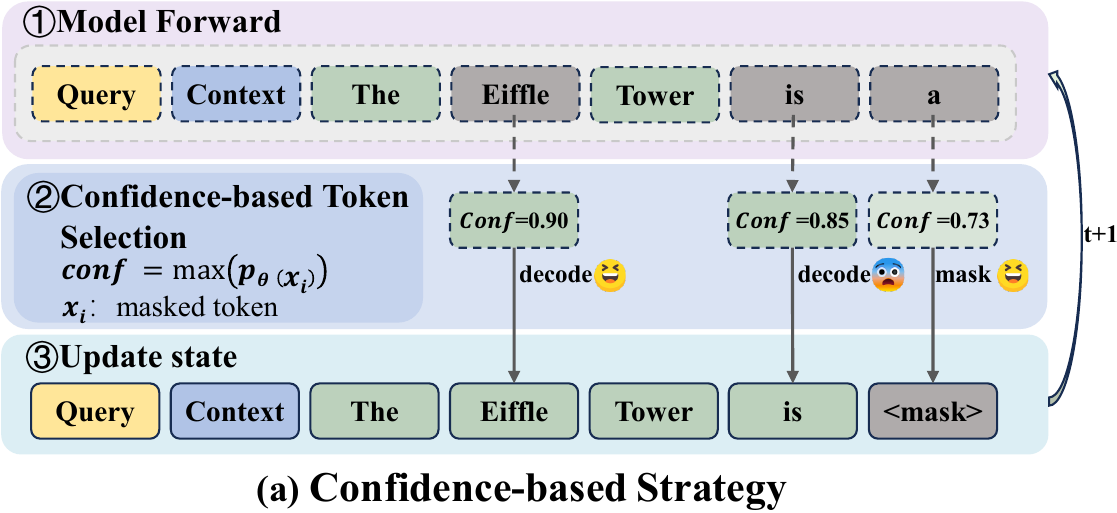} \hfill
  \includegraphics[width=\columnwidth,page=3]{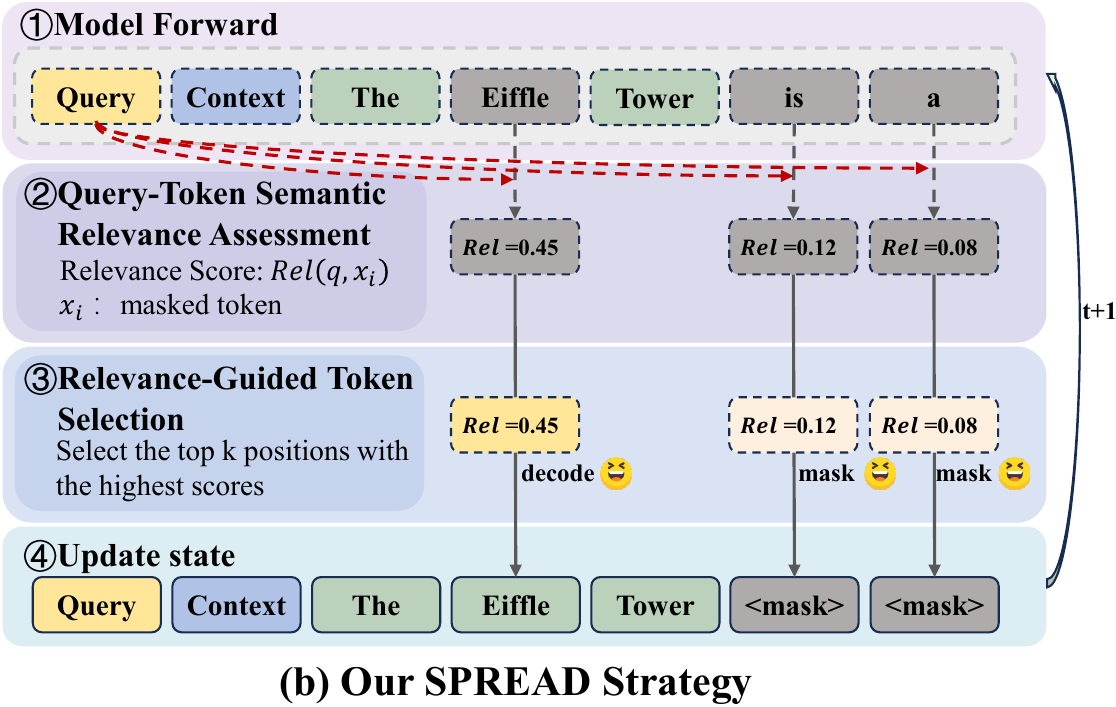}
  \caption {An overview and illustrative comparison of our proposed method: (a) Confidence-based Strategy: The existing methods rely on local model confidence for token selection, lacking guidance from the global query semantics and leading to semantic drift during generation. (b) Our \methodname~strategy: We assess the relevance of each candidate token to the query and directly guides the token selection, actively steering the generation process to maintain continuous alignment with the query semantics.}
  \label{fig:example_method}
\end{figure}

In this section, we introduce our method in detail. Our central idea is to introduce a query-relevance-guided denoising strategy at each iterative denoising step, which dynamically adjusts denoising priorities by evaluating the semantic relevance between candidate tokens at masked positions and the query, thereby guiding the generation process to consistently anchor to the query semantics.

\subsection{Overall Framework}
The framework of \methodname is illustrated in Figure~\ref{fig:example_method}. We introduce a Query-Token Semantic Relevance Assessment module at each denoising step, following the standard forward pass of the DLM.
Without requiring additional encoding overhead, this module uses the model's internal representations in the forward pass to compute a semantic relevance score between each masked position and the query. 
A Relevance-Guided Token Selection strategy then utilizes these scores to determine which token predictions should be decoded in the current step and which should remain masked for refinement in subsequent steps. 
This process is repeated in every iteration, ensuring that continuous semantic alignment is maintained throughout the entire denoising process. Next, we elaborate on these two modules.

\subsection{Query-Token Semantic Similarity Assessment} 
We realize the semantic alignment by measuring semantic similarity. Our method is based on the principle that a token's similarity to the query in the semantic space directly measures its consistency with the query's overall semantics. By computing and utilizing this semantic relevance, we can effectively steer the generation process to remain anchored to the query's semantics. To precisely measure the consistency of these local generation decisions with the global query semantics, we design a semantic relevance assessment module based on the model's internal representations.

\paragraph{Contextualized Embedding Extraction}
Given the sequence state ${x}_t$ at the denoising step $t$ and the query $q$, we utilize the model's own forward pass to obtain contextualized hidden layer representations. Let the model output hidden states for all layers be $\mathbf{H} \in \mathbb{R}^{L \times N \times d}$, where $L$ is the number of layers, $N$ is the sequence length, and $d$ is the hidden dimensionality.
We select the hidden state of the last layer  as the final contextualized representation where the representation of the $i$-th position is denoted as $\mathbf{h}_i \in \mathbb{R}^{d}$. 
Compared to static token embeddings, $\mathbf{h}_i$ incorporates deep semantic information from the sequence context, allowing it to more accurately reflect the semantic content. %
This approach requires only a single model forward pass and introduces no additional computation for token embeddings.

\paragraph{Semantic Similarity Computation} 
To maintain consistency in evaluation, the query $q$ also obtains its complete semantic representation through a forward pass of the same model. We compute the cosine similarity between the hidden state $\mathbf{h}_i$ and the query representation $\mathbf{h}_q$ as the measure of their semantic relevance:
\begin{equation}
\text{Sim}(i, q) = \frac{\mathbf{h}_i \cdot \mathbf{h}_q}{\|\mathbf{h}_i\| \|\mathbf{h}_q\|}.
\end{equation}
This metric is aligned in computational principle with the \metricname~metric defined in section~\ref{sec:metric}.

Then we map the similarity to a normalized relevance score. To smoothly amplify high-relevance signals, we apply the Sigmoid function $\sigma(\cdot)$ :
\begin{equation}
\text{Rel}(i, q) = \sigma(\text{Sim}(i, q)), 
\end{equation}
where $\text{Rel}(i, q) \in (0, 1)$ directly represents the estimated probability that the content at position $i$ is semantically aligned with the query $q$.

\subsection{Relevance-Guided Token Selection}
Based on the computed $\text{Rel}(i, q)$, \methodname~employs a straightforward and efficient strategy to guide the denoising process.

At denoising step $t$, the model produces an initial prediction distribution for all masked positions $i \in M_t$. We first calculate the relevance scores $\{ \text{Rel}(i, q) \}_{i \in M_t}$ for these positions. Subsequently, we select the top $k$ positions with the highest scores, where $k$ is a predefined or dynamically determined hyperparameter, and decode the predicted tokens at these positions, replacing the original masks. The remaining masked positions retain their masked status, waiting for further evaluation and prediction in subsequent iterative steps.

By prioritizing the retention of the token most relevant to the query, our strategy directly suppresses the accumulation of unrelated information and enforces alignment with the query's semantics, which reduces \metricname, and thereby enhances the precision of the generated answers. The pseudo-code of \methodname~is shown in Appendix~\ref{sec:algo}.

\section{Experiments}

    \begin{table*}

      \setlength{\tabcolsep}{2.0pt}
      \resizebox{\linewidth}{!}{
      \begin{tabular}{llccccccccccccc}
        \toprule
         \multicolumn{2}{c}{\multirow{2}{*}{Model}}    & \multicolumn{2}{c}{NQ} & \multicolumn{2}{c}{Trivia} & \multicolumn{2}{c} {HotpotQA}    & \multicolumn{2}{c}{MuSiQue}    & \multicolumn{2}{c}{Multihop}  & \multicolumn{2}{c}{UltraDomain} \\

                 \cmidrule[1pt](lr){3-4}\cmidrule[1pt](lr){5-6}\cmidrule[1pt](lr){7-8}\cmidrule[1pt](lr){9-10}\cmidrule[1pt](lr){11-12}\cmidrule[1pt](lr){13-14}\
                               & & $\mathrm{P}$        & $\mathrm{\metricname}$       
                                & $\mathrm{P}$        & $\mathrm{\metricname}$     
                                & $\mathrm{P}$        & $\mathrm{\metricname}$  
                                & $\mathrm{P}$        & $\mathrm{\metricname}$ 
                                & $\mathrm{P}$        & $\mathrm{\metricname}$ 
                                & $\mathrm{P}$        & $\mathrm{\metricname}$ \\
     \midrule
           \multirow{3}{*}{LLM}& \text{Qwen2.5-7B}    &   21.02        &  32.88     & 37.81     &  16.47   &  45.40    &  26.41     &   19.61    &   40.50    &   25.53     &  49.95       &    13.42    & 72.42 \\
           &\text{Qwen2.5-32B}    &   15.25        &  47.80     &   28.42   & 28.42    &  33.77    &  43.72     &   16.47    &   60.95    &  22.62      & 54.66        &   13.47     & 71.86  \\
           &\text{LLaMA3-8B }    &     14.47      &  59.21     &    32.55  & 29.23    &  35.22    &  38.45     &   13.19    &  55.78     &   12.99     &  61.79       & 18.97       & 71.37\\

        \midrule

         \multirow{6}{*}{DLM}&   \text{LLaDA (random)}     &  20.63    &  29.35    &    30.38  &    16.09  &  35.60   &  27.19     &   19.28       &   28.16      &   28.07   &  37.37      &   21.58      &  \underline{63.60} \\
                                 & \text{LLaDA (low-conf)}  &   \underline{26.05}    &  \underline{23.01}     &    \underline{36.04}   &    \underline{11.12}  &  \underline{41.38}     &  \underline{24.40}     &   \underline{23.91}        &   \underline{20.92}       &  \underline{44.91}       & \underline{24.28}      &     \underline{23.89}    &   65.57\\
                                 &  {LLaDA (\methodname) }   & \textbf{30.37}     &  \textbf{21.79}    &    \textbf{42.46}  & \textbf{8.91}    &  \textbf{48.71}    &  \textbf{19.26}     &   \textbf{26.67}       &  \textbf{17.82}       & \textbf{48.73}     &  \textbf{19.79}     & \textbf{29.33 }       & \textbf{61.48}\\

        \cmidrule(lr){2-14}
        & \text{Dream (maskgit-plus)}     &   \underline{37.11}       &  \underline{12.60}          &   \underline{47.08}      &  3.48       &   \underline{52.42}         &  6.04      &   \underline{31.35}         &   4.51         &   \underline{37.05}       & 32.13      &    36.24     &  62.62\\
                                 &  \text{Dream (entropy)}       &   35.25        &  12.61          &   44.30       &  \underline{3.44}      &   50.16       &  \underline{5.54}      &   29.33         &   \underline{4.02}         &  35.85       & \underline{32.12}     &   \underline{41.69}     & \underline{59.27}   \\
                                 &  \text{Dream (\methodname)}   & \textbf{52.79} &  \textbf{4.83} &\textbf{55.71}& \textbf{1.29}   &  \textbf{59.37} &  \textbf{4.05}   &   \textbf{37.78} &  \textbf{1.13}  &  \textbf{59.09}           &  \textbf{9.46}  & \textbf{58.19} & \textbf{24.66}\\

        \bottomrule
      \end{tabular}
      }
      \caption{The results of different methods for the question answering task. P indicates Precision ($\uparrow$) and RSD indicates Response Semantic Drift ($\downarrow$). The best results are in \textbf{bold} and the second-best results are \underline{underlined}. }
      \label{tab:main}
    \end{table*}

This section presents a systematic experimental evaluation to assess the effectiveness of our proposed \methodname~method. Our experiments are designed to answer three key research questions: 
\begin{itemize}[leftmargin=0.4cm]
    \item \textbf{Q1:} Can \methodname~consistently improve answer precision and suppress \metricname~across various datasets and model baselines?
    \item \textbf{Q2:} Does \methodname~show superior performance compared to existing DLM denoising strategies? 
    \item \textbf{Q3:} How does \methodname~affect different aspects of DLMs, such as copy rate and efficiency?%
    
\end{itemize}

\subsection{Experimental Setup}\label{exp:mainsetup}

\paragraph{Datasets} 
To comprehensively evaluate model performance across diverse knowledge retrieval and integration scenarios, we conduct experiments on six widely-used open-domain question-answering (QA) datasets: Natural Questions (NQ)~\citep{dataset-NQ} and TriviaQA~\citep{dataset-triviaqa}, which focus on factual QA, the multi-hop QA datasets HotpotQA~\citep{dataset-hotpotqa}, MuSiQue~\citep{dataset-MuSiQue}, and MultiHop-RAG~\citep{dataset-multihoprag}, which require reasoning by synthesizing information from multiple documents, and UltraDomain~\citep{dataset-ultradomain}, which challenges the model's capability for in-depth understanding within a vertical domain. These datasets encompass varying difficulties and requirements. The statistics and more details are provided in Appendix~\ref{sec:dataset}.

\paragraph{Evaluation Metrics} 
Guided by the problem analysis in Section~\ref{sec:metric}, we employ Precision (denoted as P) and \metricnamefull~(\metricname) as our core evaluation metrics, directly measuring the factual accuracy and internal semantic coherence of the the answers, respectively. Additionally, we use the Copy Rate (denoted as CR) and Redundancy (denoted as R) as auxiliary analytical metrics to gain deeper insights into the model's generation behavior, with both metrics calculated at the word level. Specifically, Copy Rate quantifies the extent of verbatim reproduction from the source context by computing the proportion of overlapping words between the generated answer and the provided context. Redundancy measures the repetitiveness within the generated answer itself, indicating an inefficient or circular information presentation.

\paragraph{Baseline Models} 
We select two representative open-source DLMs as our base models: LLaDA-8B-Instruct~\citep{llada} and Dream-7B-Instruct~\citep{ye2025dream}. For LLaDA, we compare three denoising strategies, including the native random strategy and confidence-based strategy in their original article, and our \methodname. For Dream, we compare its entropy-based strategy and maskgit-plus strategy, and \methodname.

\paragraph{Implementation Details}
We use a dense vector retriever based on the NV-Embed-v2 encoder, with document chunks of 2,000 characters, retrieving the top-5 relevant passages as context for each query. To ensure a fair comparison, all experiments, including baselines and our method, share the same generation hyper-parameters: the number of diffusion steps is 128, the maximum new token length is 512, and the sampling temperature is 0.1.
All reported results are averaged across all queries within that dataset. %

\subsection{Main Results}

To answer \textbf{Q1} and \textbf{Q2}, Table~\ref{tab:main} provides a comprehensive performance comparison between \methodname~and the baselines, evaluated on both LLaDA and Dream base models with all denoising strategies across all datasets.

The results demonstrate that \methodname~consistently and significantly improves the precision of generated answers in all experimental settings, achieving an average absolute increase of 15.34\% on LLaDA and 30.90\% on Dream. This directly shows that guiding the generation process with query semantics effectively filters out irrelevant information, leading to more focused and accurate responses. Corresponding to the gain in precision, \methodname~also significantly reduces \metricnamefull, with an average relative reduction exceeding 10.92\% on LLaDA and 61.18\% on Dream. This validates our core hypothesis that aligning the generation trajectory with the query semantics fundamentally mitigates \metricname~between sentences, thereby enhancing overall coherence.

\subsection{Further Analysis}
To address \textbf{Q3} and gain further insights into our method, we conduct a series of analyses.

\subsubsection{Impact on Copy Rate}

\begin{figure}[t]
    \includegraphics[width=\columnwidth,page=4]{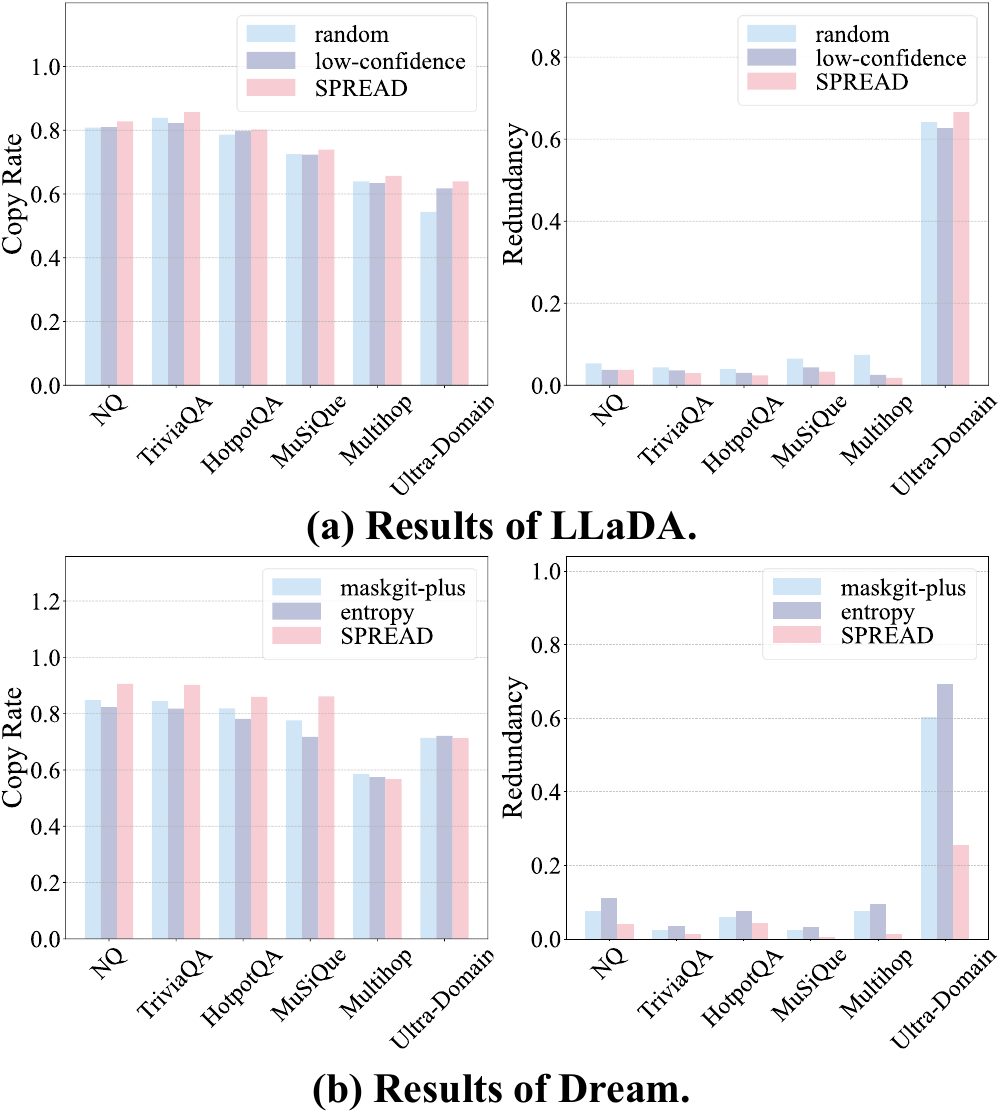}
  \caption {A comparative analysis of Copy Rate and Redundancy across different denoising strategies.}
  \label{fig:copy_dream}
  \vspace{-0.3cm}
\end{figure}

In this subsection, we analyze the model's behavior in copying content from retrieved context. As shown in Figure~\ref{fig:copy_dream}, \methodname~does not significantly alter the overall Copy Rate of the answers. Meanwhile, Redundancy is decreased significantly. The results support our argument that in RAG, how many retrieved contents are copied alone are insufficient measurements, but what content is copied is far more critical. By guiding the model to prioritize copying content strongly related to the query semantics, \methodname~significantly enhances the quality and integration of the retrieved information for enhanced question answering ability.%

\subsubsection{Multi-dimensional Analysis}
\begin{figure}[t]

  \includegraphics[width=0.49\columnwidth,page=1]{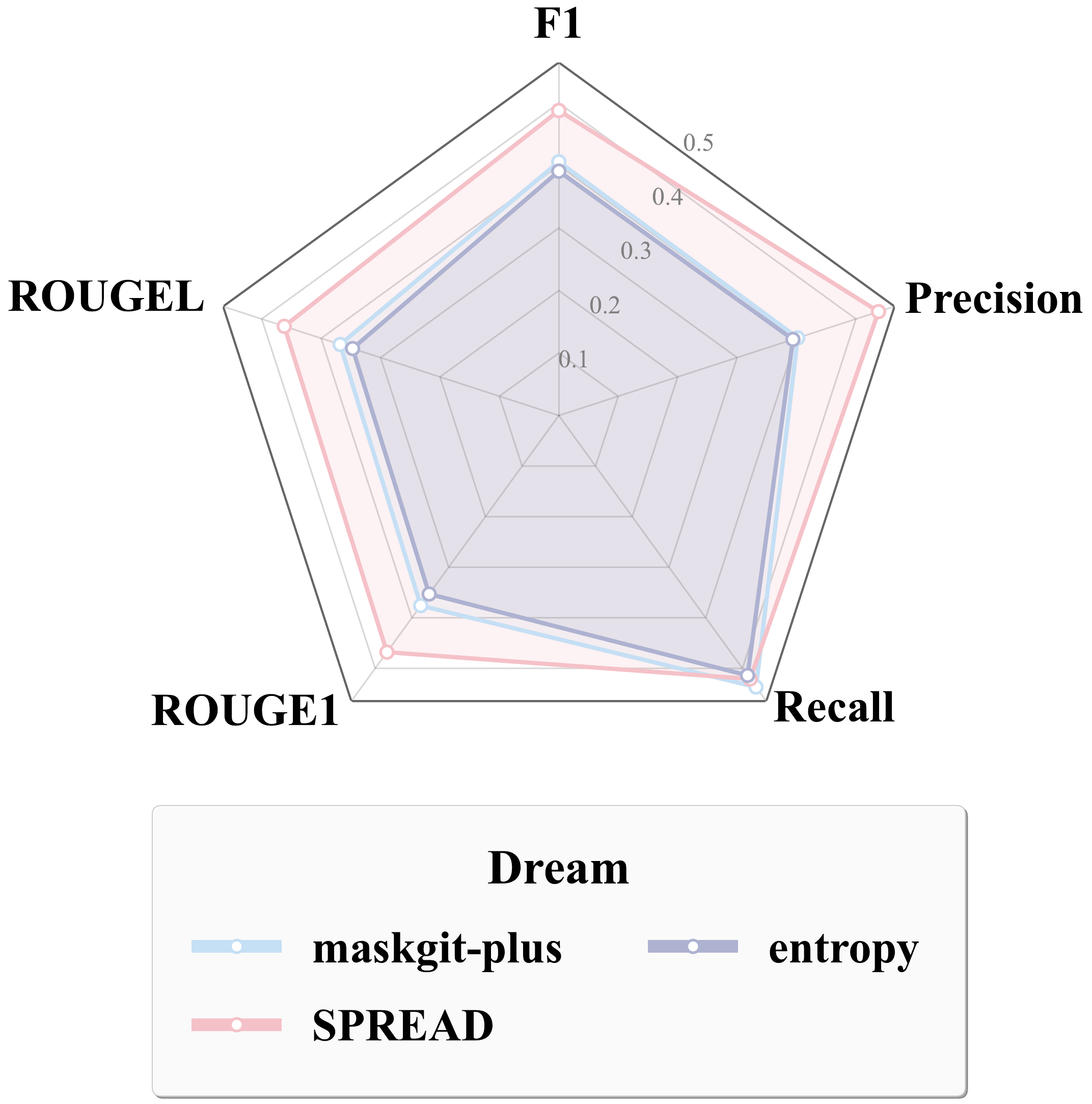} \hfill
  \includegraphics[width=0.49\columnwidth,page=1]{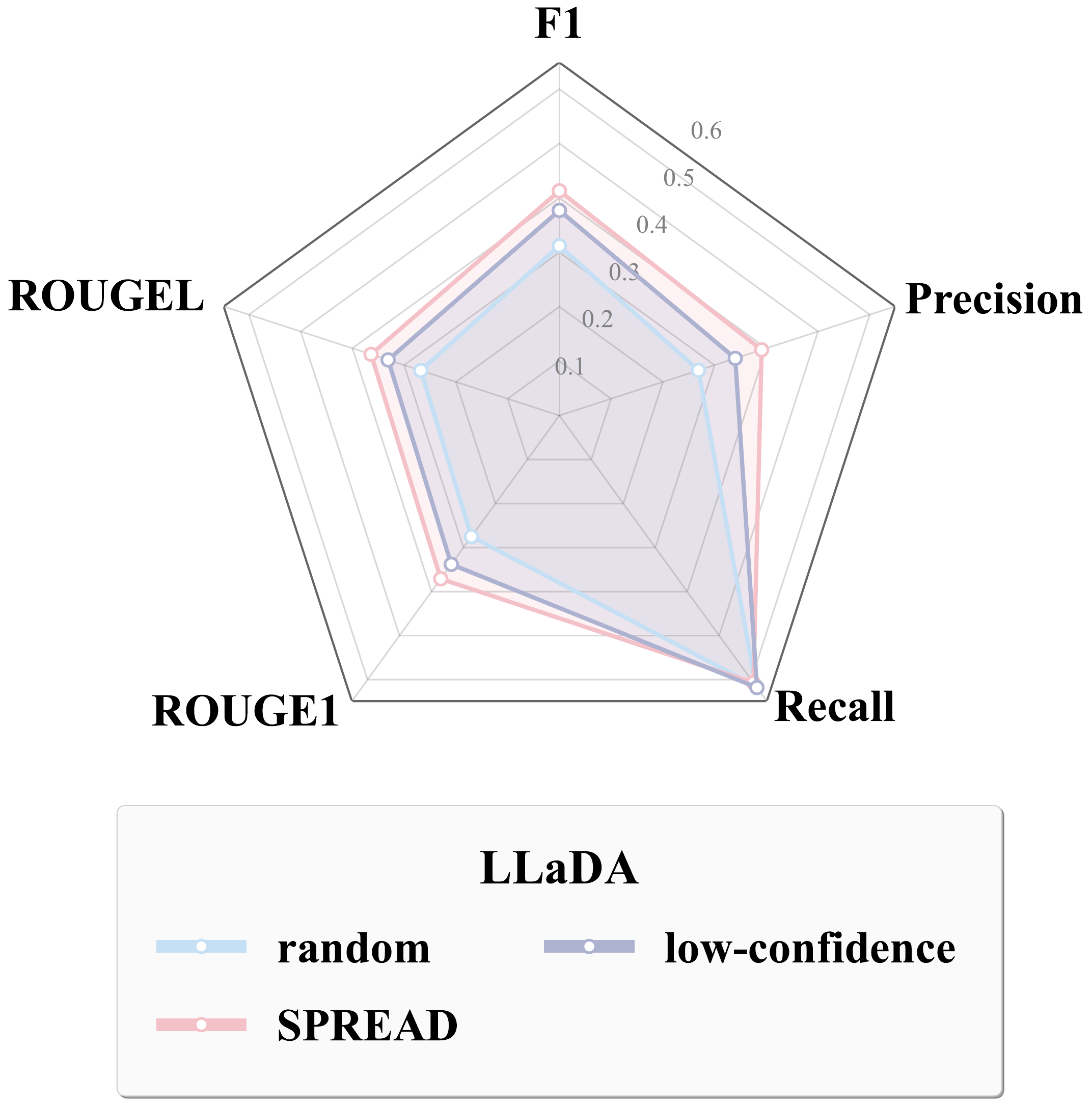}
  \caption {The results of multi-dimensional comparison of different masking strategies.  }
  \label{fig:radar}
\end{figure}

To provide an integrated view of our method’s performance across multiple evaluation dimensions, we use a radar chart to summarize five key metrics: F1, Recall, and Precision, together with ROUGE-1 and ROUGE-L which measures unigram and longest-sequence overlap against reference texts, respectively. As shown in Figure~\ref{fig:radar}, \methodname~covers a noticeably larger area than all baseline strategies, demonstrating its overall superiority. The chart highlights that our method achieves dominant performance in four of the five metrics, including F1, Precision, ROUGE-L, and ROUGE-1. This reflects an enhanced ability to generate answers that are factually accurate. %
Meanwhile, its Recall score remains on par with the strongest baselines, which indicates that our method boosts precision and coherence without sacrificing its capacity to cover relevant information from the retrieved context. This holistic evaluation confirms that \methodname~delivers a more balanced and effective generation quality, excelling in the aspects most crucial for reliable question answering tasks.

\subsubsection{Efficiency Analysis}

\begin{table}[t]

\setlength{\tabcolsep}{2.7pt}
\resizebox{\columnwidth}{!}{
    \begin{tabular}{clcc}
    \toprule
    Model & Strategy & Avg Time (s) & Tokens/sec  \\
    \midrule
    
    \multirow{3}{*}{Dream}  & maskgit-plus  & 19.64 & 30.72   \\
                            & entropy        & \textbf{18.34} & \textbf{32.41}   \\
                            & \methodname    & 19.02 & 31.49   \\
    \midrule

    \multirow{3}{*}{LLaDA}  & random          & \textbf{27.33} & \textbf{21.08} \\
                            & low-confidence & 27.39 & 21.03 \\
                            & \methodname     & 27.56 & 20.90  \\
    \bottomrule
    \end{tabular}
}
\caption{Efficiency comparison of different denoising strategies. The best results are in \textbf{bold}.}
\label{tab:efficiency_comparison}
\end{table}

To comprehensively assess the practical efficiency of our method, we conduct a detailed efficiency analysis comparing the computational cost of \methodname~against various baseline denoising strategies.
We adopt two metrics, including the Average Generation Time, measured for generating full answers on the validation set, which reflects the end-to-end cost of the iterative denoising process, and Tokens/second, which directly compares the decoding speed across different models. All experiments are conducted on a single NVIDIA H200 GPU with 141 GB of memory. 

As shown in Table~\ref{tab:efficiency_comparison}, the computational overhead introduced by \methodname~is minimal. Compared to the baseline, \methodname~incurs an increase of only 0.23s on LLaDA and 0.68s on Dream. %
This slight overhead is attributed to the computation of query-token semantic relevance scores, which is extremely efficient as it reuses the model's existing hidden states without requiring extra encoding.
  
In summary, our \methodname~strategy achieves significant improvements in answer precision and coherence by introducing only negligible computational time overhead, which ensures its practicality and efficiency for real-world applications.

\subsubsection{Relationship with Other Metrics}

\begin{table}
  \centering
  \begin{tabular}{lcc}
    \toprule
    \textbf{Metric Pair} & \textbf{LLaDA} & \textbf{Dream}\\
    \midrule
    \metricname-Precision     & -0.36  & -0.24           \\
    \metricname-Redundancy     &  0.46      & 0.55      \\
    \metricname-Copy Rate     &  -0.30  &  -0.16       \\
    \bottomrule
  \end{tabular}
  \caption{The Pearson correlation of \metricname~with Precision, Copy Rate and Redundancy. }
  \label{tab:correlation}
\end{table}

To empirically analyze the relationship between \metricname~and classical evaluation metrics in text generation, we calculate the Pearson correlation coefficients between \metricname~and Precision, Redundancy, and Copy Rate, with results presented in Table~\ref{tab:correlation}. 

The results reveal a moderate negative correlation between \metricname~and precision across both DLM backbones, statistically confirming that semantic incoherence within generated text is a plausible factor that impairs generation precision. %

Furthermore, the results show a strong positive correlation between \metricname~and redundancy, affirming their intrinsic connection in terms of redundant information. %
In other words, our metric can potentially identify generated content that is novel in wording but suffers from topic jumps or logical fractures between sentences, a blind spot for existing redundancy metrics. 

Lastly, \metricname~shows a weak negative correlation with copy rate. The result suggests that a high copy rate per se is not related to semantic drift, i.e., a model can perform ``high-quality copying'' by coherently integrating valuable information from the retrieved context and generate consistent answers. %

In sum, the empirical results further validate that \metricname~can systematically quantifies the internal semantic coherence of generated answers. 

\section{Conclusion}

This work investigates the integration of discrete DLMs within the RAG paradigm. We first identify and formalize a key challenge: \metricnamefull~(\metricname), which quantifies the progressive deviation from query semantics in generated answers. 
To address this limitation, we propose \methodname, a novel framework that steers the generation trajectory using a query-relevance-guided denoising strategy. 
Comprehensive experiments demonstrate that \methodname~delivers consistent and significant improvements by effectively enhancing answer precision and substantially reducing \metricname. 

\section{Limitations}

While \methodname~demonstrates significant efficacy, its performance is contingent on the relevance of retrieved context, while co-designing the retriever and generator can create a feedback loop for enhanced performance. Additionally, we current only validate our method on benchmark QA tasks and we aim to extend it for more open-ended generation tasks such as story writing.

\bibliography{custom}

\appendix

\begin{table*}
\centering
\vspace{-0.2cm}

\setlength{\tabcolsep}{3.3pt}
\begin{tabular}{rrrrrrr}
\toprule
 Dataset & NQ & Trivia & HotpotQA & MuSiQue & MultiHop & UltraDomain  \\
\midrule

\multirow{1}{*}{Num of Queries} & 1,000 & 1,000 & 1,000 & 1,000 & 1,000 & 958 \\
\multirow{1}{*}{Num of Passages} & 7,830 & 1,432 & 66,581 & 29,898 & 609 & 161 \\
\multirow{1}{*}{Average Length of Question} & 9.19 & 13.21 & 16.17 & 13.96 & 45.99 & 12.74 \\
\multirow{1}{*}{Average Length of Answer} & 2.61 & 2.02 & 2.44 & 2.73 & 1.30 & 39.31 \\
\bottomrule
\end{tabular}
\caption{Dataset Statistics}
\label{tab:datasets}
\end{table*}

\section{Algorithm Details}\label{sec:algo}
In this section, we provide the pseudo-code of our method in Algorithm~\ref{alg:iprad}.

\section{More Details about Datasets}\label{sec:dataset}
 Detailed dataset descriptions are as follows.
\begin{itemize}[leftmargin=0.5cm]
\item \textbf{Natural Questions}~\cite{dataset-NQ}: A large-scale question answering benchmark dataset consisting of real user queries paired with Wikipedia pages.
\item \textbf{TriviaQA}~\cite{dataset-triviaqa}: A large-scale reading comprehension dataset with complex, compositional questions and independently collected evidence documents, designed to challenge cross-sentence and multi-document reasoning.
\item \textbf{HotpotQA}~\cite{dataset-hotpotqa}: A challenging question answering dataset built on Wikipedia, containing question-answer pairs that require finding and reasoning over multiple supporting documents, with sentence-level annotated supporting facts.
\item \textbf{MuSiQue}~\cite{dataset-MuSiQue}: A systematically constructed multi-hop question answering dataset consisting of 2-4 hop questions generated via a bottom-up composition of single-hop queries.
\item \textbf{MultiHop-RAG}~\cite{dataset-multihoprag}: A benchmark dataset for retrieval-augmented generation that evaluates models on multi-hop queries requiring iterative retrieval and reasoning across multiple documents.
\item \textbf{UltraDomain}~\cite{dataset-ultradomain}: A benchmark dataset spanning diverse specialized domains, designed to evaluate models on long-context, high-level queries that require holistic understanding and multi-information synthesis.
\end{itemize}
Specifically, the MultiHop-RAG dataset is licensed under ODC-BY 1.0, MuSiQue under CC BY 4.0, and HotpotQA under CC BY-SA 4.0. The use of other established benchmarks follows standard scholarly citation and fair use conventions. We utilize all datasets strictly as benchmarks for experimental evaluation, which aligns with their intended research use. The statistics of these datasets are shown in Table~\ref{tab:datasets}.

\begin{algorithm}[t]
\caption{\methodnamefull~ (\methodname)}
\label{alg:iprad}
\begin{algorithmic}[1]
\Require Query $q$, retrieved context $c$, discrete diffusion model ${M}$, the total denoising steps $T$, tokens to retain per step $k$
\Ensure Generated answer $\mathbf{x}_0$
\State $\mathbf{x}_T \leftarrow [\text{MASK}] * L$
\State  $\mathbf{h}_q \gets {M}.\text{encode\_query}(q)$
\State Prefix $q$ and $c$ to the model input.

\For{$t = T$ \textbf{down to} $1$}
    \State // \textbf{Step 1: Obtain Contextualized Representations}
    \State $\mathbf{h} \gets {M}.\text{forward}(\mathbf{x}_t)$ 
    \State ${M}_t \gets \text{GetMaskedPositions}(\mathbf{x}_t)$

    \State // \textbf{Step 2: Compute Relevance Scores}
    \State $\mathbf{Rel} \gets \text{zeros}(|{M}_t|)$
    \For{$idx, i \in \text{enumerate}({M}_t)$}
        \State $\mathbf{Sim(i,q)} \gets \text{cosine}(\mathbf{h}_i, \mathbf{h}_q)$ 
        \State $\mathbf{Rel(i,q)} \gets \sigma(\mathbf{Sim(i,q)})$ 
    \EndFor

    \State // \textbf{Step 3: Relevance-Guided Top-$k$ Selection \& Update}
    \State $\text{TopKIndices} \gets \text{Argsort}(\mathbf{Rel(i,q)})[:k]$
    \For{$j \in {M}_t[\text{TopKIndices}]$}
        \State $\mathbf{x}_t[j] \gets {M}.\text{predict\_token}(\mathbf{x}_t, j)$ 
    \EndFor

    \State $\mathbf{x}_{t-1} \gets \mathbf{x}_t$
\EndFor

\State \Return $\mathbf{x}_0$
\end{algorithmic}
\end{algorithm}

\section{Ethical Considerations}
While our proposed method aims to enhance general language model performance and is verified in benchmark question answering tasks without any specific ethical designs, potential risks include its misuse for generating persuasive misinformation and the perpetuation of biases from the underlying semantic encoder. %
Integrating source verification and fairness testing could reduce these risks.

\section{AI Usage Statement}
During the preparation of this paper, we utilize AI writing assistants to enhance the quality of our writing. Its primary role was to assist in checking for grammatical errors, correcting spelling, improving sentence structure and polishing phrasing.%
We hereby declare that all core scientific ideas, experimental designs, data analysis, and final conclusions are the original work of the authors. 
The authors take full responsibility for the accuracy and originality of the manuscript's content.

\end{document}